\documentclass[10pt, a4paper]{article}

\usepackage{lrec-coling2024} 
\usepackage{booktabs}

\title{ReproHum \#0087-01: Human Evaluation Reproduction Report for \textit{Generating Fact Checking Explanations}\\ \vspace*{.5\baselineskip}}

\name{Tyler Loakman\textsuperscript{1}, Chenghua Lin\textsuperscript{2}} 

\address{\textsuperscript{1}Department of Computer Science, The University of Sheffield, UK\\
\textsuperscript{2}Department of Computer Science, The University of Manchester, UK \\
         tcloakman1@sheffield.ac.uk, 
         chenghua.lin@manchester.ac.uk\\}

\abstract{
This paper presents a partial reproduction of \textit{Generating Fact Checking Explanations} by \citet{atanasova-etal-2020-generating-fact} as part of the ReproHum \cite{belz:thomson:2024} element of the ReproNLP shared task to reproduce the findings of NLP research regarding human evaluation. This shared task aims to investigate the extent to which NLP as a field is becoming more or less reproducible over time. Following the instructions provided by the task organisers and the original authors, we collect relative rankings of 3 fact-checking explanations (comprising a gold standard and the outputs of 2 models) for 40 inputs on the criteria of \textit{Coverage}. The results of our reproduction and reanalysis of the original work's raw results lend support to the original findings, with similar patterns seen between the original work and our reproduction. Whilst we observe slight variation from the original results, our findings support the main conclusions drawn by the original authors pertaining to the efficacy of their proposed models. 
 \\ \newline  }

\begin{document}

\maketitleabstract

\section{Introduction}
Recently, many works have investigated the role of human evaluation in assessing the quality of outputs in the field of Natural Language Processing (NLP) and Natural Language Generation (NLG) \cite{belz-etal-2023-missing,clark-etal-2021-thats, van-der-lee-etal-2019-best}. Whilst human evaluation is often seen as the gold standard method of evaluation which takes into account the perceptions of real human end-users, there is much debate over the reproducibility of such evaluation \cite{belz-etal-2023-missing,howcroft-etal-2020-twenty}. Automatic metrics, whilst scalable, frequently demonstrate poor concurrent validity, correlating poorly with human judgements \cite{liu2024llms,zhao-etal-2023-evaluating,alva-manchego-etal-2021-un,reiter-2018-structured, belz-reiter-2006-comparing}. However, the performance of human evaluation has likewise been shown to have multiple flaws, including ill-defined evaluation criteria compounded by the absence of sufficient evaluator/annotator training to attenuate the subjectivity of the texts being rated from the subjective interpretation of the evaluation criteria itself. Furthermore, several works have discussed the presence of poorly selected human panels, including sufficient language proficiency and task understanding \citet{schoch-etal-2020-problem}. This is further hindered by the choice of many works to obfuscate these shortcomings by neglecting to report any demographic information regarding participants, including for highly subjective language types such as humour \cite{loakman-etal-2023-iron}. Such discrepancies have resulted in widespread troubles in reproducing the results of different works in NLP \cite{10.1162/coli_a_00508}.

It is for reasons such as these that the ReproHum shared task aims to shine a spotlight on the level of reproducibility within the field of NLP through the mass reproduction of contemporary research through its many partner labs so that poor practices are identified and a record can be made of the progress of reproducibility over time, as researchers become increasingly aware of the best practices to follow in performing human evaluation in their works.

\section{Background}
As participants in the ReproHum project, we selected the paper \textit{Generating Fact Checking Explanations} by \citet{atanasova-etal-2020-generating-fact} as the focus of our reproduction, owing to interest in the topic of explanation generation, and previous experience of being part of evaluator panels for similar research. Through the automatic selection process, the ReproHum team identified the single experiment and criterion that we were to attempt to reproduce the results from, as introduced in $\S$\ref{sec:reproduction}.

Owing to our participation in the ReproHum project \cite{belz:thomson:2024}, we were provided with the following materials: (i) a guide to the common approach to reproduction, (ii) the original paper and dataset required to perform a reproduction, and (iii) additional documents pertaining to clarifications and additional information provided by the original authors once contacted. During this process, the authors of this paper (and therefore the team performing the reproduction) did not contact the authors of the original work directly at any stage.

In performing this reproduction, we adhered to the following criteria outlined in the documentation provided by the ReproHum organisers. All participants were paid minimally to the UK National Living Wage (12GBP per hour) as set by the ReproHum team for pair pay, in which we specifically paid 15GBP for this task and paid via Amazon Vouchers from our estimation that the task would take approximately 1.25hrs (which was confirmed by our evaluators following completion). Additionally, this work underwent ethical review and approval by the ethics review board of the primary author's institution (where all participants in this reproduction were also selected).

\section{Original Study}
\label{sec:original}

In recent years with the widespread sharing of misinformation and the coining of "fake news", the need for accurate and reliable fact-checking systems has grown exponentially. While existing systems have demonstrated impressive performance, their "black box" nature often obscures the reasoning behind their predictions. This lack of transparency can hinder user trust and limit the adoption of these systems. \citet{atanasova-etal-2020-generating-fact} identified an overall research focus on the veracity prediction task of news claims in existing research and a lack of work focusing on generating natural language explanations to justify these veracity predictions. They aimed to address the main drawback of a black-box system by generating explanations to support the assigned veracity labels. To do this, the authors leverage detailed fact-checking reports (termed "ruling comments") published alongside veracity labels by fact-checking organisations to produce explanations that resemble human-written justifications. This approach is further bolstered through a multi-task learning framework, where explanation generation is jointly optimised with a veracity prediction task for a DistilBERT \cite{sanh2020distilbert} based model. This joint training enables the system to identify regions in the ruling comments that are not only close to the gold standard explanation but also contribute to the overall fact-checking decision.

\subsection{Evaluation}
\label{sec:orig_eval}

The authors evaluate their approach using both automatic and human evaluation methods. While automatic evaluation relies on the standard metric of ROUGE \cite{lin-2004-rouge}, human evaluation focuses on a range of different criteria listed below, alongside their original definitions:

\begin{itemize}
    \item \textbf{\underline{Coverage}} - The explanation contains important, salient information and does not miss any important points that contribute to the fact check.
    \item \textbf{Non-redundancy} - The summary does not contain any information that is redundant/repeated/not relevant to the claim and the fact check.
    \item \textbf{Non-contradiction} - The summary does not contain any pieces of information that are contradictory to the claim and the fact check.
\end{itemize}

Based on these criteria, evaluators are requested to rank different explanations based on their performance on each criterion (as well as providing an \textit{Overall} ranking).
The original results in \citet{atanasova-etal-2020-generating-fact} demonstrate that the multi-task learning approach leads to improved performance for both veracity prediction and explanation generation. Notably, the generated explanations achieve better coverage and overall quality compared to explanations trained solely to mimic human justifications. This suggests that the joint training framework allows the system to capture the knowledge required for accurate fact-checking, leading to more informative and relevant explanations. In our reproduction, we focus solely on the underlined criterion of \textit{Coverage}.

\section{Reproduction Setting}
\label{sec:reproduction}
\paragraph{Task Setting}

\begin{figure*}
\centering
    \includegraphics[width=0.75\linewidth]{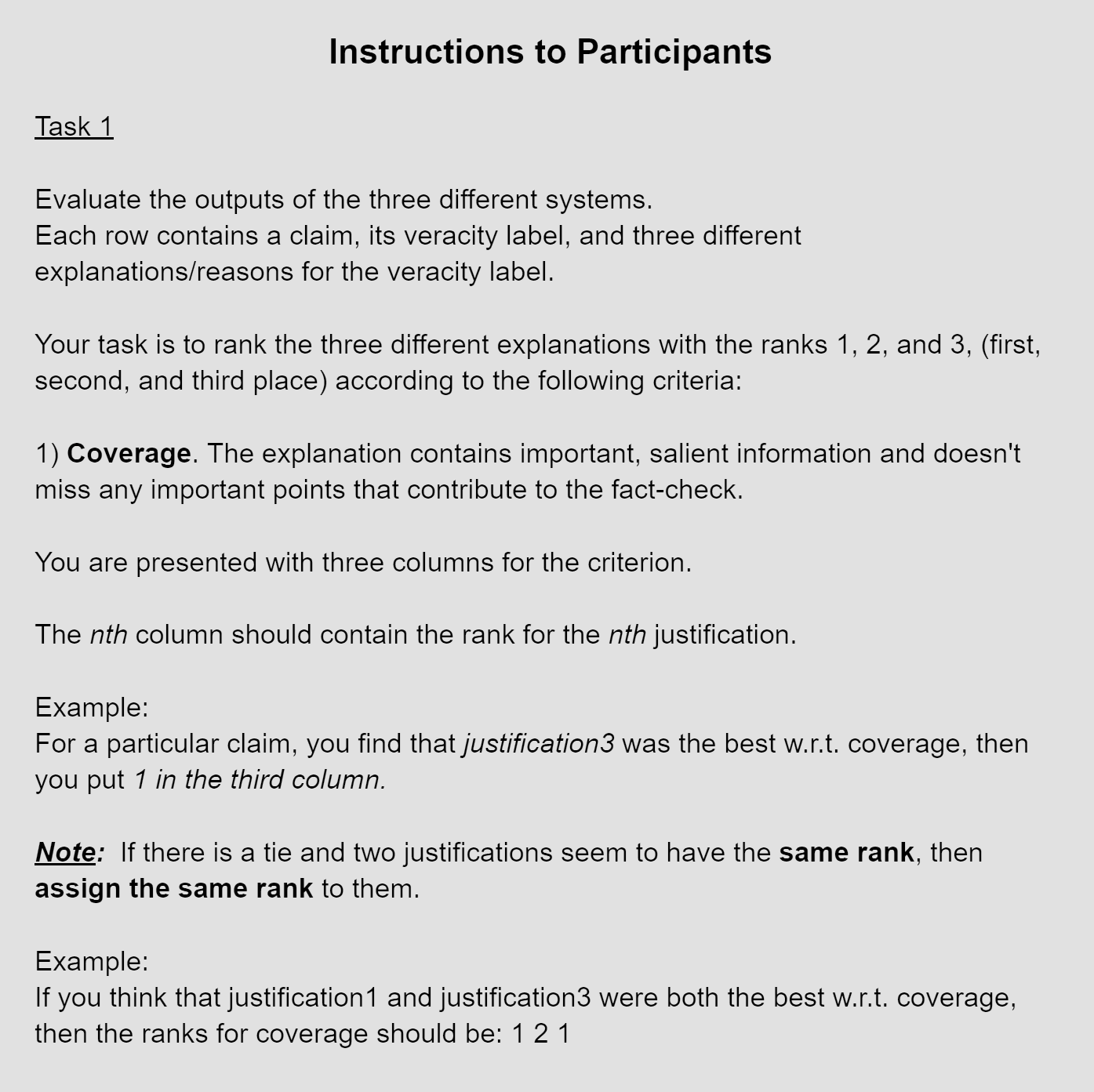}
    \caption{Modified instructions from \citet{atanasova-etal-2020-generating-fact} presented to participants within the reproduction. we made minor modifications to the original instructions presented to participants in order to remove information related to tasks and criteria that were not to be assessed in this reproduction.}
    \label{fig:instructions}
\end{figure*}

As directed by the ReproHum team, we performed our reproduction on a single element of the original work by \citet{atanasova-etal-2020-generating-fact} regarding evaluating outputs on the aforementioned criteria of \textit{Coverage}.
We presented the same instructions to participants as presented by \citet{atanasova-etal-2020-generating-fact} with minor changes, as presented in \autoref{fig:instructions}. These changes exclusively involve the removal of information regarding other evaluation criteria used in the original study outside of \textit{Coverage}, including \textit{Non-redundancy}, \textit{Non-contradiction}, and a holistic \textit{Overall} rating. We additionally remove all mention of the separate Task 2 which is not the subject of this reproduction. As with the original study, we performed our reproduction experiment by having participants place their relative preference rankings of 3 systems (i.e., a gold standard and two models) in a spreadsheet facilitated via Google Sheets. Within this, 3 columns follow the 3 explanations (from the 3 different models) to place rankings (where the $n$-th column contains the ranking for the $n$-th justification), as outlined in \autoref{fig:instructions}. In line with the recommended approach to performing reproductions presented by the ReproHum team, we additionally incorporate data validation techniques in the form of drop-down boxes containing rankings of 1-3. This ensured that participants only entered valid options in the ranking task. We present model outputs to participants in the same shuffled order presented in the original paper to also avoid order effects and bias towards particular columns. In total, each participant annotated 120 items, consisting of the outputs of 3 systems (including the human gold standard) for 40 inputs. We also make available a HEDS datasheet \cite{shimorina-belz-2022-human} detailing the process of our reproduction study.\footnote{Available at \url{https://github.com/nlp-heds/repronlp2024}.}

\paragraph{Evaluator Demographics}
In the original work by \citet{atanasova-etal-2020-generating-fact} we have limited demographic details regarding the participants. However, we are aware that they are colleagues of the authors and have experience in fact-checking annotation tasks, whilst not exclusively being native speakers of the target language. In our replication, we use 3 Ph.D. students in Natural Language Processing, all of which have experience in fact-checking and related tasks (e.g., misinformation/rumour detection). All participants in our reproduction also have a professional working level of English fluency.

\section{Results}

\begin{table*}[ht]
    \centering
    \small
    \begin{tabular}{cccc}
        \toprule
        \multicolumn{4}{c}{\textbf{Original}} \\
        \midrule
        \textbf{Annotators} & \textbf{Gold} & \textbf{Explain-Extr} & \textbf{Explain-MT} \\
        \midrule
        All & \textbf{1.48} & 1.89 & \underline{1.68} \\
        1\textsuperscript{st} & \textbf{1.50} & 2.08 & \underline{1.87} \\
        2\textsuperscript{nd} & \textbf{1.74} & 2.16 & \underline{1.84} \\
        3\textsuperscript{rd} & \textbf{1.21} & 1.42 & \underline{1.34} \\
        \midrule
        CV* & 6.52\% & 7.12\% & 8.52\%\\
        \midrule
        \multicolumn{4}{c}{\textbf{Recreation}} \\
        \midrule
        \textbf{Annotators} & \textbf{Gold} & \textbf{Explain-Extr} & \textbf{Explain-MT} \\
        \midrule
        All & \textbf{1.47} & 1.88 &  \underline{1.69} \\
        1\textsuperscript{st} & \textbf{1.48} & 2.08 & \underline{1.90} \\
        2\textsuperscript{nd} & \textbf{1.72} & 2.15 &  \underline{1.87} \\
        3\textsuperscript{rd} & \textbf{1.21}  & 1.41 & \underline{1.33} \\
        \midrule
        CV* & 7.19\% & 7.65\% & 7.93\%\\
        \midrule
        \multicolumn{4}{c}{\textbf{Reproduction}} \\
        \midrule
        \textbf{Annotators} & \textbf{Gold} & \textbf{Explain-Extr} & \textbf{Explain-MT} \\
        \midrule
        All & \textbf{1.58} & 2.03 & \underline{1.83} \\
        1\textsuperscript{st} & \textbf{1.43} & 2.43 & \underline{2.08} \\
        2\textsuperscript{nd} & \underline{1.68} & 1.80 & \textbf{1.55} \\
        3\textsuperscript{rd} & \textbf{1.65} & 1.88 & \underline{1.85} \\
        \bottomrule
    \end{tabular}
    \caption{Comparison between \citet{atanasova-etal-2020-generating-fact} and our reproduction on the criterion of "Coverage". Values present the Mean Average Ranks (MAR) of the explanations. The explanations come from the gold justification (\textbf{Gold}), the generated explanation (\textbf{Explain-Extr}), and the explanation learned jointly (\textbf{Explain-MT}) with the veracity prediction model. A lower MAR indicates a better average ranking. For each row, the best results are in \textbf{bold}, and the best automatically generated explanations are \underline{underlined}. "Annotators" refers to each individual rater, whilst "All" is the mean across all annotators. \textit{CV*} refers to the Coefficient of Variation for the mean ratings of the 3 systems compared to our reproduction results following the implementation by \citet{belz-2022-metrological}. \textit{Original} refers to the results presented in the original paper by \citet{atanasova-etal-2020-generating-fact}, whilst \textit{Recreation} refers to the results we gain by reanalysing the original study's data exclusively for the same sample that our evaluators were presented. Finally, \textit{Reproduction} refers to the results of our reproduction study using our new evaluators. The ordering of annotators across \textit{Recreation} and \textit{Original} should be considered arbitrary, as we cannot fully guarantee each line corresponds to the same annotator as the original.}
    \label{tab:results}
\end{table*}

We present the results of the original study and our reproduction in \autoref{tab:results}. Due to minor discrepancies in the specific evaluated materials (owing to some evaluators in the original work assessing approximately 80 items, and others assessing only 39, with some omissions), we additionally report what we term a "recreation", where we reanalyse the original paper's raw data to facilitate a direct comparison against only the same 40 inputs as presented to our evaluators. In the original work by \citet{atanasova-etal-2020-generating-fact}, the criterion of \textit{Coverage} is shown to have low inter-annotator agreement as calculated via Krippendorff's Alpha \cite{Krippendorff2019}, reporting $\alpha = 0.26$ across their 3 evaluators. In our reproduction, we find slightly better agreement among our participants, with $\alpha =  0.35$ when specifically accounting for an ordinal level of measurement, whilst we find agreement across the 40 evaluated inputs in the original data to be very similar to what was reported for the particular subset used by the authors in the original work ($\alpha = .27$)

In terms of overall patterns seen in the data, the results of our reproduction can be seen to differ slightly from those of the original. Firstly, in the original study, the golden human-authored explanations were preferred by all participants, whilst this is not seen to be the case in our reproduction or in our reanalysis of a specific subset of the original paper's raw data (i.e., \textit{recreation}) Instead, we find only 2 of our 3 participants to rank the golden explanations in their expected 1\textsuperscript{st} place. However, in terms of the automatically generated explanations we observe the \textit{Explain-MT} model (where the explanation is learnt jointly with the veracity prediction model) to outperform \textit{Explain-Extr} (where the auxiliary veracity prediction model is learnt separately), mirroring the results presented in the original work. 

Furthermore, when aggregating the results of all 3 evaluators in our reproduction, we can see that the overall rankings assigned to each output are higher (i.e., worse) than the findings of \citet{atanasova-etal-2020-generating-fact}. However, whilst our raw figures differ from the original findings (owing to the relatively subjective task criteria and small evaluator panel sizes), our findings reflect the same overall patterns as the original work, with the human-authored golden explanations \textit{Gold} outperforming the authors' proposed models in the majority of cases, whilst the more complex \textit{Explain-MT} model, which is trained alongside a veracity prediction task, outperforms the \textit{Explain-Extr} model that learns to generate explanations in isolation.

To compare against the original study's findings, we calculate correlations between our results and those provided by the original paper's authors using Spearman's $\rho$ and Pearson's $r$. Due to the original work's raw data having results for more than 40 trials, and with some missing values, we assess only the same 40 trials as presented to our participants (equivalent to the \textit{Recreation} in \autoref{tab:results}) and calculate the mean rank given to each output by the evaluators (which is robust to cases where not all evaluators in the original work assessed a given output). The results show a strong correlation between the results of our reproduction and the original study ($\rho = .524$ and $r = .541$, which are both significant at $\alpha = .01$), demonstrating that we were able to reproduce the general evaluator preferences observed in the original experiment.

\section{Conclusion}

In this paper, we have presented our reproduction findings for an element of human evaluation presented in \citet{atanasova-etal-2020-generating-fact} regarding the criteria of \textit{Coverage} to compare gold standard fact-checking explanations with 2 proposed models. In terms of overall comparison with the original work, we find a higher level of rating agreement among our evaluator panel than demonstrated in the original work but also observe a slightly different overall pattern than presented by the original authors, with one of the proposed models ranking higher than the gold standard human-authored explanation from 1 of our 3 participants. We do, however, observe the same pattern when reanalysing the raw data from the original study, focussing exclusively on the same subset of examples presented to our evaluators in the reproduction. Additionally, our reproduction lends credence to the results presented by \citet{atanasova-etal-2020-generating-fact} regarding the model trained to generate explanations alongside a veracity prediction model (Explain-MT) outperforming the model that is trained to generate explanations in isolation (Explain-Extr) in terms of human rankings.
It is important to note, however, that the result of our reproduction covers only one of the multiple human evaluation criteria on which the raters were asked to assess the generations in the original work, and this pattern may not necessarily be present across all different criteria.

Overall, we reiterate the importance of performing reproduction studies such as this in order to assess the trend of reproducibility within the field of NLP. Within this paper, we have successfully reproduced the findings of the original work with some minor variability (likely owing to the small size of the evaluation panels in the original work, and consequently our reproduction). This is particularly salient for the topic of generating fact-checking explanations that \citet{atanasova-etal-2020-generating-fact} tackle, as this constitutes a high-impact application of NLP with an increased need for reliable and robust models and evaluation procedures in order to avoid the effects of misinformation.

\section{Acknowledgements}
We would like to thank the organisers of the ReproHum/ReproNLP shared task for their efforts in bringing this large-scale reproduction effort to light and for their continued assistance throughout the process of performing our reproduction experiments. We would also like to thank the authors of \textit{Generating Fact Checking Explanations} \cite{atanasova-etal-2020-generating-fact} for their transparency in providing the necessary resources and materials to aid in our reproduction of their work.

Tyler Loakman is supported by the Centre for Doctoral Training in Speech and Language Technologies (SLT) and their Applications, funded by UK Research and Innovation [grant number EP/S023062/1]

\nocite{*}
\section{Bibliographical References}\label{sec:reference}

\bibliographystyle{lrec-coling2024-natbib}
\bibliography{lrec-coling2024-example}


\end{document}